\documentclass[letterpaper]{article}
\usepackage[preprint]{aaai2027}
\usepackage[hyphens]{url}
\usepackage{graphicx}
\usepackage{natbib}
\usepackage{caption}
\usepackage{algorithm}
\usepackage{algorithmic}
\usepackage{booktabs}
\usepackage{amsfonts}
\usepackage{amsmath}
\usepackage{amssymb}
\usepackage{multirow}

\newcommand{\method}{\textsc{NS}}

\newcommand{\tfv}{\textsc{TFv6}}
\newcommand{\carla}{\textsc{CARLA}}

\title{Herding End-to-End Autonomous Driving via Neuro-Symbolic Safety Guards}

\author{
    Sim\'on Pati\~no Idarraga\textsuperscript{\rm 1},
    Erick Silva\textsuperscript{\rm 2},
    Rehana Yasmin\textsuperscript{\rm 2},
    Ali Shoker\textsuperscript{\rm 2}
}
\affiliations{
    \textsuperscript{\rm 1}Universidad de Antioquia, Medell\'in, Colombia\\
    \textsuperscript{\rm 2}King Abdullah University of Science and Technology (KAUST), Thuwal, Saudi Arabia
}

\begin{document}

\maketitle

\begin{abstract}

Modern end-to-end driving agents can achieve high average performance yet still violate basic traffic rules that a human driver would never miss. The reason is structural: they learn statistical patterns rather than the physical conditions that guarantee safe driving, leaving their decision-making process opaque and safety constraints unenforced. We introduce a \textbf{\emph{neuro-symbolic safety guard}}, a lightweight module that attaches to the final command interface of an already-trained agent. Immediately before a command reaches the vehicle, it checks the command against explicit safety rules and, only when necessary, replaces it with the nearest safe alternative. Each intervention is directly executable and traceable to the rule that triggered it, while the guard itself requires no retraining and adds no learned component. Evaluated on the long-tail benchmarks \textit{Fail2Drive} and \textit{Bench2Drive} using the state-of-the-art \textit{TransFuser~v6} (\tfv) as a case study, the guard improves Success Rate by $\mathbf{15\%}$ and reduces safety-critical collisions by up to $\mathbf{53\%}$, while preserving the original Driving Score.

\end{abstract}

\section{Introduction}
\label{sec:intro}
A modern driving agent can earn a strong benchmark score and \emph{still} run a red light, brake too late behind a lead vehicle, or enter a pedestrian crossing unsafely. For autonomous driving, these rare failures matter more than average competence: they are the sparse but high-consequence events that decide whether a system is trusted at deployment, and each one erodes public confidence, intensifies regulatory scrutiny, and slows adoption \citep{zhang2024,koopman2024,alpamayo2025}. The cost is not hypothetical: the 2023 Cruise robotaxi pedestrian-dragging incident shows how a single mishandled interaction can stall an entire deployment program. Even the strongest learned agents remain brittle exactly here.

The core problem we address is that these agents fail on situations a human would find obvious. Studies of end-to-end driving models show that they often rely on statistical patterns in their training data rather than the basic traffic facts that actually govern safety, hence, competence on familiar routes does not carry over to rare ones \citep{jaeger2023tfpp,gerstenecker2026fail2drive}. Crucially, the safety requirement violated in these cases is not hidden or subtle: it can be stated as a simple physical condition on how the vehicle may move. What is missing is therefore not just a better learned model, but a way to enforce such conditions on the commands the car finally executes.
\begin{figure*}[t]
    \centering
    \includegraphics[width=0.8\textwidth]{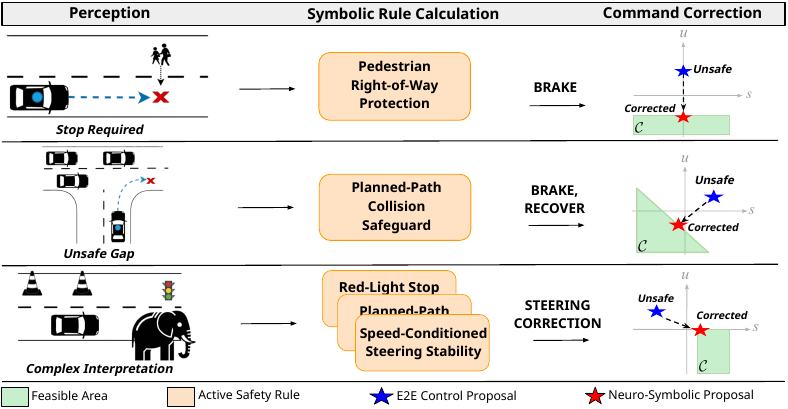}
    \caption{How the safety guard works, in three steps (left to right). Perception: the scene is read and a risky situation is identified. Symbolic Rule Calculation: the relevant traffic-safety rules are activated. Command Correction: the guard adjusts the driving command. In each right-hand plot the two axes are the actual controls sent to the car, the longitudinal command $u$ (throttle/brake) and the steering command $s$; the shaded region is the set of commands the active rules allow (green, $\mathcal{C}$). The driving agent's original command (blue, \emph{Unsafe}) falls outside this safe region, therefore the guard moves it to the nearest allowed command (red, \emph{Corrected}).}
    \label{fig:intro_overview}
\end{figure*}

We address this gap with a \emph{neuro-symbolic safety guard}: a module that is attaches to the final command interface of a trained driving agent (Figure~\ref{fig:intro_overview}). The learned agent proposes its driving command, but before that command reaches the vehicle the guard checks it against a set of explicit \emph{traffic-safety rules} and, only when the proposal is unsafe, replaces it with the closest safe alternative. Because nothing is retrained and no learned component is added, the same guard attaches to \emph{any} end-to-end model that exposes its scene signals, leaving the agent itself entirely intact. We instantiate it on a state-of-the-art agent as a case study, but the guard itself is architecture-independent.

Moreover, the \emph{traffic-safety rules} are not ad hoc: \emph{each} is derived in closed form from an established safety or vehicle-dynamics model and reduces to a single bound on throttle, brake, or steering, with the derivations given in Section~\ref{sec:method}. Because every correction is triggered by an explicit rule, one can always identify why the guard intervened, making it deterministic, modular, and auditable.

We present three major contributions in this paper:
\begin{enumerate}
\item \textbf{A neuro-symbolic safety guard for end-to-end driving.} We attach the guard to the command interface of a frozen neural agent, where it operates in three stages: it \emph{reads} the scene signals the network already exposes, \emph{reasons} over them with explicit rules, and \emph{restricts} the command to the safe set they define. Both systems then contribute at execution time, the network supplying learned flexibility and the rules the guarantees it cannot state. Unlike safe-RL shielding, which constrains actions during training, or rule-based planners, which replace the controller outright, the guard leaves the policy intact.

\item \textbf{A set of grounded traffic-safety rules.} We derive five rules in closed form from established models, responsibility-sensitive safety for longitudinal margins and the kinematic bicycle model for cornering, each reduced to one bound on throttle, brake, or steering. Every correction is therefore verifiable against a physical criterion rather than a tuned heuristic.

\item \textbf{Measured robustness and safety gains.} We compare the same frozen agent with and without the guard on the Fail2Drive long tail. Task success rises by 15\%, safety-critical collisions fall by up to 53\%, and the competence lost on unfamiliar scenes is nearly halved, at an unchanged Driving Score and without retraining.

\end{enumerate}

The remainder of the paper positions the guard against related work (Section~\ref{sec:related}), derives the traffic-safety rules and the guard that enforces them (Section~\ref{sec:method}), and reports the evaluation (Section~\ref{sec:exp}).

\section{Related Work}
\label{sec:related}

\paragraph{End-to-end driving and closed-loop benchmarks.}
End-to-end driving has progressed from conditional imitation learning \citep{codevilla2018} through trajectory-guided methods \citep{tcp2022,patcp2025} to multi-modal transformer stacks \citep{mmfn2022,interfuser2022,transfuser2021,nguyen2026lead}.
Evaluation has shifted in parallel from aggregate route scores to targeted safety and generalization analysis \citep{bench2drive2024,safebench2022,jaeger2023tfpp,gerstenecker2026fail2drive}.
The unresolved challenge is that benchmark gains do not readily translate to reliable behavior in rare, high-consequence failures.

\paragraph{Constrained action correction.}
Safe-reinforcement-learning methods project actions onto feasible sets during training \citep{shielding2018,dalal2018,optlayer2018}, and convex-optimization layers provide the theoretical substrate \citep{optnet2017,dcol2019}.
Rule-based driving planners such as PDM-Lite replace the learned controller entirely \citep{pdm2024}.
Our guard differs on both counts: it corrects only the final action of a fully trained policy, leaving the perception-and-planning backbone frozen.

\paragraph{Neuro-symbolic methods.}
Neuro-symbolic approaches couple neural perception with symbolic reasoning \citep{sarker2021,bhuyan2024}.
The taxonomy of \citet{kautz2022}, reproduced in both surveys, distinguishes patterns by where the two components meet: the Neuro$\rightarrow$Symbolic pattern is a cascade from a neural system into a symbolic reasoner, in which a neural module first produces a proposal and a downstream symbolic module then constrains or interprets it.
Most driving instantiations embed symbolic knowledge during training \citep{sharifi2026,albilani2024}.
We instead deploy the Neuro$\rightarrow$Symbolic pattern as an execution-time safety guard over a frozen policy, evaluated directly under closed-loop distribution shift.

\section{Method}
\label{sec:method}

\begin{figure*}[t]
    \centering
    \includegraphics[width=0.82\textwidth]{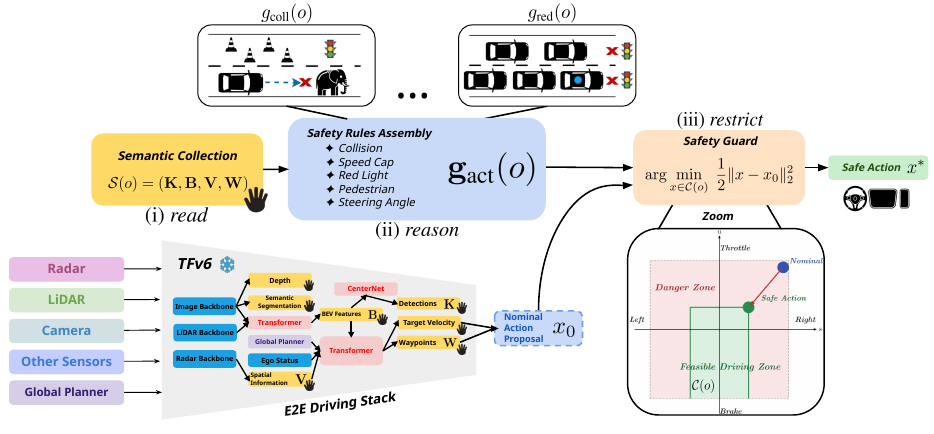}
    \caption{The neuro-symbolic safety guard in three stages, wrapping a frozen end-to-end stack (bottom) that emits its nominal command $x_0$ as usual. \textbf{(i) Read:} the same forward pass exposes a structured scene $\mathcal{S}(o)=(\mathbf{K},\mathbf{B},\mathbf{V},\mathbf{W})$, the detections, BEV features, target velocity, and waypoints already computed inside the network. \textbf{(ii) Reason:} the safety rules read this scene and each returns one limit on a control axis, assembled into the bound vector $\mathbf{g}_\text{act}(o)$. \textbf{(iii) Restrict:} these bounds define the feasible region $\mathcal{C}(o)$ onto which the guard projects $x_0$, solving $\operatorname*{argmin}_{x \in \mathcal{C}(o)} \frac{1}{2}\|x - x_0\|_2^2$ for the executed command $x^*$. Only the final command is constrained; the agent is never retrained.}
    \label{fig:architecture}
\end{figure*}

Our method inserts a \emph{neuro-symbolic safety guard} between a frozen end-to-end driving stack and the final actuation interface (Figure~\ref{fig:architecture}). The learned policy still performs perception, planning, and nominal control; the guard does not choose routes or replace the planner. Its role is narrower: it prevents the vehicle from executing a command that would move it toward an unrecoverable and dangerous state.

The guard has three parts, and this section is organized around them: the \textbf{setup} fixing notation for its two inputs (\S\ref{sec:formulation}); the \textbf{safety rules} turning a driving situation into concrete limits on the command (\S\ref{sec:constraints}); and the \textbf{guard} itself, which replaces a proposal breaking any active rule with the closest command satisfying all of them, and changes nothing otherwise (\S\ref{sec:qp}).

\subsection{Setup: What the Guard Reads}
\label{sec:formulation}

\paragraph{Control space.}
We represent the executed vehicle command as a two-dimensional control vector,
\begin{equation}
    x = [u,\, s]^\top, \qquad u = \tau - \beta, \qquad x \in \mathcal{U} = [-1,1]^2,
    \label{eq:u_encoding}
\end{equation}
where $u$ is the signed longitudinal command, $s$ is the steering command, $\tau \in [0,1]$ is throttle, $\beta \in [0,1]$ is brake, and $\mathcal{U}$ is the admissible control set; positive $u$ accelerates and negative $u$ brakes. Throttle and brake are mutually exclusive on a road vehicle, therefore, collapsing them into the single signed scalar $u=\tau-\beta$ loses no actuation authority while turning every longitudinal safety condition into one upper bound on $u$.

\paragraph{Neural backbone and nominal proposal.}
Let $\mathcal{O}$ denote the observation space and $o \in \mathcal{O}$ a single multi-sensor observation (the synchronized camera, LiDAR, and radar inputs at one time step). The backbone is an end-to-end driving policy $\pi_\theta : \mathcal{O} \rightarrow \mathcal{U}$ with learned parameters $\theta$, mapping the observation $o$ to a nominal control proposal
\begin{equation}
    x_0 = \pi_\theta(o) = [\,u_0,\; s_0\,]^\top,
    \label{eq:nominal}
\end{equation}
where $x_0$ is the command the learned stack would execute without intervention. We reuse this policy unchanged: it already solves perception, prediction, and planning at competitive accuracy, and the guard needs only its output and a few intermediate signals. In our experiments $\pi_\theta$ is \tfv{}, a multi-modal imitation-learning agent that fuses camera, LiDAR, and radar through a transformer backbone and predicts both the waypoint path and the target velocity \citep{nguyen2026lead}.

\paragraph{Structured scene state.}
The guard needs to see the scene, and modern end-to-end stacks already show it: to stay trainable and debuggable, most expose interpretable outputs in the same inference pass that yields the command \citep{interfuser2022,patcp2025,alpamayo2025}. We reuse four of them (the scene block feeding the guard in Figure~\ref{fig:architecture}),
\begin{equation}
    \mathcal{S}(o) = (\mathbf{K}, \mathbf{B}, \mathbf{V}, \mathbf{W}),
    \label{eq:scene_state}
\end{equation}
CenterNet-style object detections $\mathbf{K}$ (position, class, confidence \citep{zhou2019centernet}), a bird's-eye-view (BEV) semantic map $\mathbf{B}$, radar returns $\mathbf{V}$ (radial velocity and range), and the policy's own planned waypoints $\mathbf{W}$. Each encodes traffic facts the policy computes but never enforces on its command. Because no single signal is complete, we read each only for what it measures reliably and fuse rather than substitute: detections give class but not velocity, radar gives velocity but not class, and the two together give both.

This dependence on exposed signals costs us little: where a stack lacks them, the heads can be added without touching the policy, and post-incident analysis of real robotaxi deployments argues for exactly this kind of conservative, inspectable decision-making. But exposing the scene is not enforcing it, and that step is our contribution. With the scene in symbolic form, we turn to the rules that read it (\S\ref{sec:constraints}) and the guard that acts on it (\S\ref{sec:qp}).

\subsection{The Traffic-Safety Rules}
\label{sec:constraints}

This subsection presents the safety rules. A rule is not a single equation but a procedure: it takes an established safety principle, applies it to the scene signals of \S\ref{sec:formulation}, and reduces the result to one number, a limit on one control axis. We use five such rules, one per entry of the bound vector as follows:
\begin{equation}
    \mathbf{g}_\text{act}(o)
    =
    \bigl[
    \underbrace{g_\text{coll}(o)}_{\text{R1}},\,
    \underbrace{g_\text{speed}(o)}_{\text{R2}},\,
    \underbrace{g_\text{red}(o)}_{\text{R3}},\,
    \underbrace{g_\text{ped}(o)}_{\text{R4}},\,
    \underbrace{g_\text{lat}(o)}_{\text{R5}}
\bigr]^\top
    \label{eq:active_bounds}
\end{equation}
This vector \emph{is} the rule set, written numerically: each $g_\bullet(o)$ is the limit that rule imposes in the current scene. The four longitudinal limits share the interval $[u_\text{min},u_\text{max}]=[-1,1]$ of the command $u$ they constrain ($1$ full throttle, $-1$ full brake); the fifth ($g_\text{lat}$) limits steering. Table~\ref{tab:rules} states when each rule fires and how it changes the command; assembling this vector is the \emph{reason} stage of Figure~\ref{fig:architecture}, and the guard (\S\ref{sec:qp}) enforces all five at once.

\begin{table}[t]
\centering
\small
\begin{tabular}{@{}p{2.35cm}p{4.85cm}@{}}
\toprule
\textbf{Rule (bound)} & \textbf{Description} \\
\midrule
\multicolumn{2}{@{}l}{\textit{Longitudinal rules (limit $u$)}} \\
\textbf{R1. Planned-Path Collision Safeguard} \newline ($g_\text{coll}$) & Prevents the vehicle from following its intended waypoint path into a solid obstacle. The rule first suppresses forward acceleration and escalates to braking when the remaining gap becomes unsafe. \\
\textbf{R2. Speed-Limit Compliance} \newline ($g_\text{speed}$) & Keeps the vehicle inside the urban speed envelope by reducing positive longitudinal command and, if necessary, requesting mild braking until the speed returns to a safe legal range. \\
\textbf{R3. Red-Light Stop Compliance} \newline ($g_\text{red}$) & Forces the command toward a full stop when a red light is active and the remaining distance no longer supports safe continuation through the intersection. \\
\textbf{R4. Pedestrian Right-of-Way Protection} \newline ($g_\text{ped}$) & Enforces early yielding whenever a pedestrian occupies, or is about to enter, the forward crossing corridor in front of the ego vehicle. \\
\midrule
\multicolumn{2}{@{}l}{\textit{Lateral rule (limits $s$)}} \\
\textbf{R5. Speed-Conditioned Steering Stability} \newline ($g_\text{lat}$) & Shrinks the admissible steering range as speed increases, keeping the executed command within a safe lateral-acceleration envelope. \\
\bottomrule
\end{tabular}
\caption{The five safety rules. Each row is one rule: the entry it fills in the bound vector $\mathbf{g}_\text{act}(o)$ of Eq.~\eqref{eq:active_bounds}, the driving situation it covers, and the correction it can impose. The first four constrain longitudinal motion ($u$); the last constrains steering ($s$).}
\label{tab:rules}
\end{table}

Together these five span the dominant ways an urban command turns unsafe: an obstacle on the planned path, excess speed, a red light, a pedestrian, and loss of control in a fast turn. They share one construction principle: ground an established physical or safety model directly in the commands the car executes. The same procedure therefore extends cleanly to hazards beyond these five. R1 turns the Responsibility-Sensitive Safety stopping distance \citep{shalev2017rss} into a throttle-and-brake bound; R5 turns the bicycle model of vehicle dynamics into a steering bound that holds before the tires lose grip.

\paragraph{Planned-Path Collision Safeguard (R1, COLL).}
COLL is the central rule, preventing the vehicle from driving into obstacles on the path the policy intends to follow: vehicles, stopped obstacles, oncoming traffic, and out-of-distribution objects such as animals or debris. Its defining choice is to be \emph{path-relative} rather than box-relative, asking whether an obstacle lies on the planned trajectory rather than inside a fixed window ahead. A fixed forward box brakes needlessly for parked cars at the kerb and for oncoming traffic on curves; gating on the planned path removes these false positives while still catching obstacles the ego is driving toward. COLL computes its bound in four steps.

\emph{Step 1 (planned path).} We form the path polyline from the predicted waypoints ($\mathbf{W}$) by TFv6,
\begin{equation}
    \mathcal{P}(o) = \bigl[(0,0),\, \mathbf{w}_1,\, \ldots,\, \mathbf{w}_T\bigr],
    \label{eq:path}
\end{equation}
in ego coordinates, where $\mathbf{W} = [\mathbf{w}_1,\ldots,\mathbf{w}_T]^\top$ and $T$ is the prediction horizon.

\emph{Step 2 (on-path gate).} An obstacle at position $\mathbf{p}$ is treated as a threat only if it lies within a corridor around this path,
\begin{equation}
    d\!\left(\mathbf{p},\, \mathcal{P}(o)\right) \leq \omega(o),
    \label{eq:path_gate}
\end{equation}
where $d(\cdot,\mathcal{P}(o))$ is point-to-polyline distance and $\omega(o)$ is the corridor half-width. The corridor is wide enough to cover the intended path; when the waypoints show a sustained lateral excursion, that is, the policy is planning around the obstacle, it tightens, and the guard defers to that evasive maneuver instead of blocking it indefinitely.

\emph{Step 3 (two perception channels).} Two complementary detectors share the gate, following the fuse-not-substitute discipline of \S\ref{sec:formulation}. Channel~A is class-aware, starting from camera detections and using radar only to fuse longitudinal velocity onto an associated object. Channel~B uses radar as a class-agnostic solid-object detector, catching out-of-distribution hazards no learned class covers. The shared path gate is what makes a class-agnostic sensor safe to brake on, since off-path clutter is filtered as in Channel~A. COLL takes the more conservative bound.

\emph{Step 4 (safe distance to bound).} For the gated obstacle we require an RSS-style stopping margin \citep{failsafe2021},
\begin{equation}
    d_\text{safe} = \begin{cases}
        \dfrac{v_\text{ego}^2}{2\,a_\text{ego}} - \dfrac{v_\text{obs}^2}{2\,a_\text{obs}} + v_\text{ego}\,\delta_\text{brake},
        & v_\text{obs} \ge 0, \\[1.5ex]
        \dfrac{v_\text{closing}^2}{2\,a_\text{ego}} + v_\text{closing}\,\delta_\text{brake},
        & v_\text{obs} < 0,
    \end{cases}
    \label{eq:coll_dsafe}
\end{equation}
where $v_\text{ego}$ and $v_\text{obs}$ are ego and obstacle speeds, $v_\text{closing}=v_\text{ego}-v_\text{obs}$ is closing speed, $a_\text{ego}$ and $a_\text{obs}$ are conservative braking rates, and $\delta_\text{brake}$ is the actuation delay. Both cases are physical: when the obstacle moves with traffic ($v_\text{obs}\ge0$) the ego must cover its own braking distance less the distance the obstacle clears, plus a reaction term; when it approaches ($v_\text{obs}<0$) only closing speed matters, as its motion cannot be credited. Converting $d_\text{safe}$ into the largest speed still permitting braking within the remaining gap gives $g_\text{coll}(o)$; a close-range emergency cap dominates when the gap becomes critical, and positive throttle is blocked whenever an on-path obstacle is present and not clearly pulling away.

\paragraph{The remaining four rules.}
The other four rules (R2--R5) share a single pattern: convert a scene-measured distance into the fastest speed from which the car can still stop, then clip the command to it. For a hazard at buffered range $d$ under braking $a_\text{brake}$, the induced longitudinal bound is
\begin{align}
    v_\text{stop} &= \sqrt{2\,a_\text{brake}\,\max(d,\,0)}, \nonumber\\
    g_\bullet(o) &= \operatorname{clip}\!\left(\frac{v_\text{stop} - v_\text{ref}}{v_\text{comfort}},\, u_\text{min},\, u_\text{max}\right). \label{eq:template}
\end{align}
Here $v_\text{stop}$ is the \emph{safe speed}: the highest speed from which the car can still brake to a full stop within the available distance $d$, obtained from the constant-deceleration stopping relation $v^2=2\,a_\text{brake}\,d$. The bound then measures the current speed $v_\text{ref}$ against this budget: when $v_\text{ref}$ exceeds $v_\text{stop}$ the term is negative and the command is braked, and when it is below the rule stays slack; $v_\text{comfort}$ normalizes this speed error into the control range $[u_\text{min},u_\text{max}]$.

The four rules instantiate this template through their choice of $d$ and $v_\text{ref}$, plus one rule-specific guard (Table~\ref{tab:rules}). Red-light (R3) is representative: $d$ is the buffered distance to the stop line, $v_\text{ref}=v_\text{ego}$, and the bound is weighted by the cue confidence $P_{\text{eff},\text{tl}}$ (BEV-semantic primary, detector fallback); a low-confidence light yields a graded restriction rather than an abrupt stop. The others follow with the natural substitutions: worst-case closing speed for a pedestrian, the urban ceiling for the speed limit, and for R5 the bicycle-model cap $s_\text{max}=\max\!\big(A_{\text{lat,max}}L/(g_s v_\text{ego}^2),\, s_\text{floor}\big)$, binding only when a raw command would imply loss of control.

\subsection{The Safety Guard}
\label{sec:qp}

We can now state the guard compactly. At each time step it maps the policy's proposal and the current scene to a corrected command $x^{*}$ in three moves, the \emph{read}, \emph{reason}, and \emph{restrict} stages of Figure~\ref{fig:architecture}:
\begin{equation}
    \begin{aligned}
        \text{(i) \emph{read}:}\quad & x_0 = \pi_\theta(o), \\
        \text{(ii) \emph{reason}:}\quad & g(o) = \mathcal{R}\big(\mathcal{S}(o)\big), \\
        \text{(iii) \emph{restrict}:}\quad & x^{*} = \Pi_{\mathcal{C}(o)}(x_0),
    \end{aligned}
    \label{eq:three_phase}
\end{equation}
where the policy proposes $x_0$ (\S\ref{sec:formulation}), the rules ($\mathcal{R}$) turn the scene $\mathcal{S}(o)$ into the bound vector $g(o)$ (\S\ref{sec:constraints}), and $\Pi$ projects the proposal onto the feasible set $\mathcal{C}(o)$ the rules define. The rest of this subsection makes the ``restrict'' step precise. It solves a scene-dependent correction problem: given the nominal proposal $x_0$ and the current observation $o$, return the closest admissible command $x^*$. The observation defines the feasible control set
\begin{equation}
    \mathcal{C}(o) \,=\, \{x \in \mathcal{U} : \mathbf{D}x \leq g(o)\},
    \label{eq:feasible_set}
\end{equation}
where the command is the two-axis control $x=[u,s]^\top\in\mathcal{U}$ (longitudinal $u$, steering $s$), $g(o)$ is the scene-conditioned bound vector assembled from the active rules and actuator limits, and $\mathbf{D}$ is the fixed constraint-structure matrix that maps each active rule to the control axis it limits. Each rule thus contributes one affine constraint, and together they define the feasible set. The executed command is the Euclidean projection of the nominal proposal $x_0$ onto this set, i.e., the quadratic program (QP)
\begin{equation}
    x^* = \operatorname*{argmin}_{x \in \mathcal{C}(o)} \frac{1}{2}\|x - x_0\|_2^2.
    \label{eq:qp}
\end{equation}
When several rules over-constrain the command, the most restrictive admissible bound wins, resolving the conflict in favor of safety.

\subsection{Why the Guard Can Be Trusted}
\label{sec:algorithm}

Algorithm~\ref{alg:filter} instantiates the three stages of Eq.~\eqref{eq:three_phase} on the single backbone forward pass that produces $x_0$, adding only lightweight, bounded computation that runs at control rate.

\begin{algorithm}[t]
\caption{Neuro-Symbolic Safety Guard (one time step)}
\label{alg:filter}
\begin{algorithmic}[1]
\REQUIRE Observation $o$; nominal proposal $x_0 = [u_0,\, s_0]^\top$ from $\pi_\theta(o)$; ego speed $v_\text{ego}$
\ENSURE Executed command $x^{*} = [u^{*}, s^{*}]^\top$
\STATE \textbf{Extract} the structured scene state $(\mathbf{K}, \mathbf{B}, \mathbf{V}, \mathbf{W})$ from the same forward pass that produced $x_0$
\STATE Build the planned-path polyline $\mathcal{P}(o)$ from $\mathbf{W}$ and test whether the waypoints indicate bypass intent
\STATE Compute the planned-path collision bound $g_\text{coll}$ from the path-gated camera and radar channels; update the short-horizon coast / recovery logic if a confirmed obstacle briefly disappears
\STATE Compute the speed-limit, red-light, and pedestrian bounds $g_\text{speed}$, $g_\text{red}$, $g_\text{ped}$
\STATE Compute the steering-stability bound $g_\text{lat} \leftarrow s_\text{max}(v_\text{ego})$
\STATE Assemble the scene-conditioned bound vector $g(o)$ from these rule outputs and the actuator limits
\STATE \textbf{Solve} the projection (QP) $x^{*} \leftarrow \arg\min_{x \in \mathcal{C}(o)} \tfrac{1}{2}\|x - x_0\|_2^2$, with $\mathcal{C}(o)=\{x\in\mathcal{U}:\mathbf{D}x\le g(o)\}$
\RETURN $x^{*}$
\end{algorithmic}
\end{algorithm}

The guard earns trust from how it is built rather than from tuning. \emph{It preserves good driving:} because $x^{*}$ is the closest admissible command to $x_0$, the guard moves as little as the active rules require and leaves an already-safe proposal untouched. \emph{Its corrections are safe by construction:} every longitudinal bound is a brake-feasible speed derived from scene geometry (Eqs.~\eqref{eq:coll_dsafe},~\eqref{eq:template}), preserving a kinematically feasible stopping condition rather than a tuned threshold; this holds conditional on perception reporting the hazard correctly, a limit we quantify in \S\ref{sec:exp_results}. \emph{And every correction is auditable:} a given scene always yields the same correction (Algorithm~\ref{alg:filter}), traceable to the single rule that caused it.

\section{Experiments}
\label{sec:exp}

\subsection{Evaluation Design}
\label{sec:eval_design}
To separate real driving skill from pattern-matching on familiar scenes, we re-stage the same hazard with different assets and layouts (a \emph{distribution shift}): a capable policy still copes, whereas an overfit one degrades, with unsafe decisions concentrated in these shifted scenes. This motivates two hypotheses. \textbf{(H1)}~On the generalization split, the guard reduces safety-critical failures (collisions with vehicles, pedestrians, and static obstacles). \textbf{(H2)}~It achieves this while preserving task competence, without trading safety for mobility. We deliberately do \emph{not} measure perception quality, ride comfort, or sample efficiency: the guard trains no weights and does not touch perception, hence we hold it fixed and vary \emph{only} the final action interface, keeping the comparison a clean attribution to the guard.

\textbf{Benchmark.} We evaluate on Fail2Drive~\citep{gerstenecker2026fail2drive}, a \carla{} v2 benchmark of 200 short routes in Town13 (mean length $219$\,m) covering 17 rare-hazard scenario classes. Its routes come in 100 matched pairs, and this pairing defines the two splits used throughout the paper. The \textbf{in-distribution split} stages each hazard with the familiar objects and layouts an agent encounters in training. The \textbf{generalization split} stages the \emph{same} hazard on the same road, changing only the appearance and arrangement of the objects involved; the situation is unfamiliar while the driving task is identical. Because the pair differs in nothing else, the drop from the first split to the second measures how much of an agent's competence was memorized rather than learned. This suits our claim better than aggregate suites such as Bench2Drive (which we also run as a competence check) or the standard \carla{} Leaderboard, which reward average competence on familiar routes rather than safety under rare shifts, and better than open-loop datasets, which cannot expose closed-loop failures at all.

\textbf{Baselines.} Our primary comparison is paired: the \emph{same} \tfv{} policy with and without the guard (denoted \textit{\method{}}, our neuro-symbolic safety guard) on the same released weights, under Fail2Drive's fixed protocol, ensuring the measured change is attributable to the guard rather than to a different model or training run. We additionally report the public Fail2Drive leaderboard~\citep{f2dleaderboard} (Table~\ref{tab:fail2drive_leaderboard}) to confirm that the backbone is a genuinely state-of-the-art starting point rather than a weak strawman. As a secondary check, we run the same paired comparison on Bench2Drive, reported briefly in \S\ref{sec:exp_results} as it only confirms preserved competence.

\textbf{Metrics.} The evaluation rests on three metrics, kept identical to the benchmark's for direct comparability. \emph{Driving Score} (DS) measures route progress, scaled down by every infraction; \emph{Success Rate} (SR) is the stricter fraction of routes completed cleanly. Their \emph{Harmonic Mean}, $\mathrm{HM}=2\,\mathrm{DS}\cdot\mathrm{SR}/(\mathrm{DS}+\mathrm{SR})$, is our primary criterion: it follows the \emph{weaker} of the two, and cannot be lifted either by rushing through routes while colliding or by crawling too cautiously to arrive. Two further analyses complete the evaluation: the in-distribution-to-shift generalization drop (which H1 and H2 target) and a per-infraction breakdown that separates a prudent stop from a dangerous contact (Section~\ref{sec:discussion}).

\textbf{Setting.} All variants run in closed loop in the open-source \carla{} simulator~\citep{carla2017} (0.9.15, Leaderboard~2.0) on a shared GPU cluster, in inference only: we train and fine-tune nothing, leaving backbone weights identical across conditions and the guard as the only moving part. We follow Fail2Drive's fixed protocol without modification (three seeds over the 100 paired routes of each split), leaving neither variant tuned to the benchmark.

\subsection{Results and Analysis}
\label{sec:exp_results}
\label{sec:discussion}
\begin{table}[t]
    \centering
    \small
    \setlength{\tabcolsep}{4.2pt}
    \begin{tabular}{@{}lcccc@{}}
        \toprule
        \textbf{Method} & \textbf{DS}$\uparrow$ & \textbf{SR}$\uparrow$ & \textbf{HM}$\uparrow$ & \textbf{$\Delta$HM}$\uparrow$ \\
        \midrule
        \multicolumn{5}{@{}l}{\emph{Learned sensor-based agents}} \\
        AlignDrive~\citeyearpar{wu2026aligndrive} & 68.6 & 62.7 & 65.5 & $-12.9$ \\
        TF++~\citeyearpar{jaeger2023tfpp} & 75.4 & 61.1 & 67.5 & $-16.5$ \\
        \tfv{}~\citeyearpar{nguyen2026lead} & 79.5 & 70.7 & 74.8 & $-18.4$ \\
        BevAD~\citeyearpar{holtz2026bevad} & \textbf{82.3} & 68.7 & 74.9 & $-12.2$ \\
        BridgeDrive~\citeyearpar{liu2026bridgedrive} & 81.9 & 75.0 & 78.3 & $-16.1$ \\
        \tfv{} + \method{} (ours) & 80.0 & \textbf{81.3} & \textbf{80.6} & \textbf{$-9.8$} \\
        \midrule
        \multicolumn{5}{@{}l}{\emph{Privileged-state / expert agents (ceiling)}} \\
        PlanT~2.0~\citeyearpar{gerstenecker2025plant2} & 73.3 & 58.0 & 64.8 & $-25.0$ \\
        PDM-Lite-F2D~\citeyearpar{gerstenecker2026fail2drive} & 94.0 & 95.3 & 94.6 & $-1.8$ \\
        \bottomrule
    \end{tabular}
    \caption{Fail2Drive generalization split: leading entries, ordered by HM. Rows cite the paper introducing each method; all baseline scores are as recorded on the public leaderboard under the benchmark's fixed protocol. $\Delta$HM is the relative HM drop from the in-distribution split, i.e.\ the competence an agent loses when the scene becomes unfamiliar; less negative is better. \textbf{Bold} marks the best \emph{learned} entry per column; privileged agents are a reference ceiling, not competitors.}
    \label{tab:fail2drive_leaderboard}
\end{table}
\begin{table}[t]
    \centering
    \small
    \setlength{\tabcolsep}{3pt}
    \begin{tabular}{@{}lccc@{\hspace{7pt}}ccc@{}}
        \toprule
        & \multicolumn{3}{c@{\hspace{7pt}}}{\textit{In-Distribution}} & \multicolumn{3}{c}{\textit{Generalization}} \\
        \cmidrule(r{5pt}){2-4}\cmidrule{5-7}
        & \textbf{DS} & \textbf{SR} & \textbf{HM} & \textbf{DS} & \textbf{SR} & \textbf{HM} \\
        \midrule
        \tfv{} & 90.2 & 93.3 & 91.7 & 79.5 & 70.7 & 74.8 \\
        \tfv{}\,+\,\method{} & 86.7 & 92.3 & 89.4 & \textbf{80.0} & \textbf{81.3} & \textbf{80.6} \\
        \midrule
        $\Delta$ (\%) & $-3.9$ & $-1.1$ & $-2.5$ & $+0.6$ & $\mathbf{+15.0}$ & $\mathbf{+7.8}$ \\
        \bottomrule
    \end{tabular}
    \caption{Paired comparison of the \emph{same} frozen \tfv{} policy without and with the guard, on both splits. The \tfv{} row is its public-leaderboard entry; \tfv{}\,+\,\method{} is our run on identical weights; every difference is the guard's. The $\Delta$ row is the \emph{relative} change the guard produces within each split. \textbf{Bold} marks improvement on the generalization split.}
    \label{tab:fail2drive_pair}
\end{table}

\begin{figure}[t]
  \centering
  \includegraphics[width=1.0\linewidth]{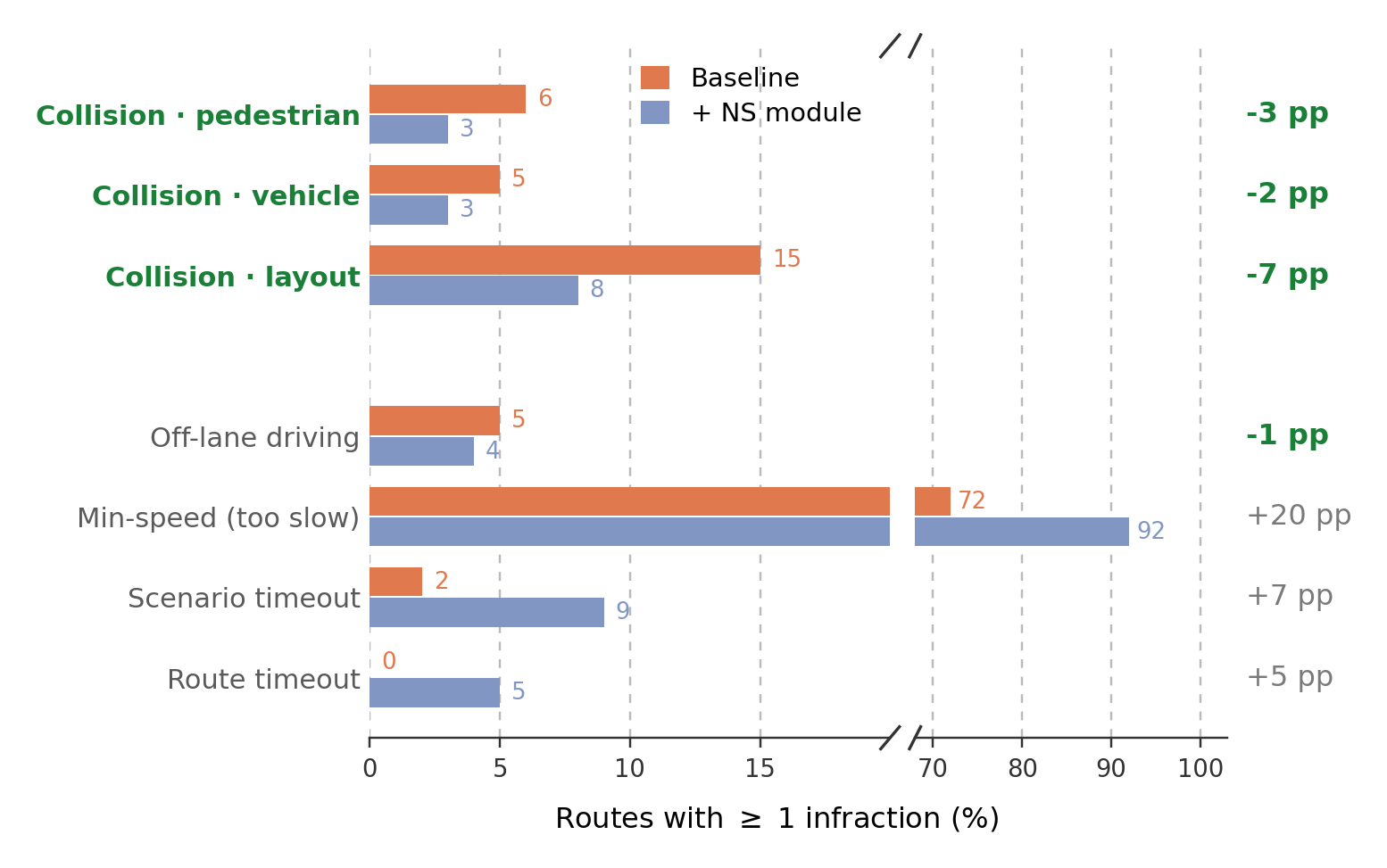}
  \caption{Per-infraction comparison (\mbox{TFv6} vs.\ \mbox{TFv6\,+\,NS}).}
  \label{fig:infraction_bars}
\end{figure}

Table~\ref{tab:fail2drive_leaderboard} places the guarded policy against the field, and Table~\ref{tab:fail2drive_pair} isolates what the guard alone changes. Read together, they point in opposite directions, and that contrast is the main result. On the in-distribution split the guard \emph{costs} $2.5\%$ HM; on the generalization split it \emph{gains} $7.8\%$. A layer that helped on both splits would simply be a better policy, and one that hurt on both would be a worse one. Helping only where the driving is unfamiliar is what a safety constraint should do: it acts when the policy is about to err and stays inactive when the policy is right.

\noindent\emph{Hypothesis 1 (\textbf{H1}): safety improves where it matters.} On the generalization split, the guard lifts every metric of the frozen policy it wraps: SR by $10.6$ points ($+15.0\%$) and HM by $5.8$ ($+7.8\%$), without lowering Driving Score (Table~\ref{tab:fail2drive_pair}). More informative than the scores themselves is how much each agent \emph{loses} between the two splits, the $\Delta$HM column of Table~\ref{tab:fail2drive_leaderboard}. Fail2Drive reports an average loss of $16.3\%$ across the models it surveys; \tfv{} alone loses $18.4\%$, and the guard reduces this to $9.8\%$, the smallest loss of any learned agent and second only to a privileged expert that reads ground-truth simulator state. A small loss is easy to obtain by scoring poorly on both splits, and must be read together with the starting score: AlignDrive and BevAD lose less than \tfv{} ($-12.9\%$ and $-12.2\%$), but start at in-distribution HM of $75.2$ and $85.3$ against \tfv{}'s $91.7$. The guarded policy is the only entry that starts high and loses little.

\noindent\emph{Hypothesis 2 (\textbf{H2}): nominal competence is largely preserved.} On the in-distribution split, the guard costs $3.5$ DS, $1.0$ SR and $2.3$ HM (Table~\ref{tab:fail2drive_pair}). This cost is expected, and its direction is fixed by the design: the guard can only narrow the set of allowed commands, never widen it (\S\ref{sec:qp}). Where the policy already drives well no rule should fire, and a rule that fires anyway can only remove progress the policy would have made safely. This split therefore measures how often the guard intervenes unnecessarily, and $2.5\%$ HM is the price of the conservative safety margins in Eq.~\eqref{eq:coll_dsafe}. Bench2Drive shows the same pattern on scenarios the agent knows well (DS $94.5\!\rightarrow\!89$, SR $99.0\!\rightarrow\!94.8\%$), indicating that the cost follows how often hazards occur rather than anything specific to Fail2Drive.

\noindent\emph{What the guard trades.} Figure~\ref{fig:infraction_bars} shows the mechanism behind both splits. Collisions fall in every category: pedestrian collisions from $6.3\%$ of routes to $3.0\%$, layout collisions from $15.0\%$ to $7.7\%$, and vehicle collisions from $4.7\%$ to $3.3\%$. What increases instead is lost time: min-speed infractions from $72.3\%$ to $92.0\%$, and timeouts by $5$--$8$ points. Because the guard can only slow the car, every intervention is paid for in delay. This is also why DS and SR move so differently ($+0.6\%$ against $+15.0\%$): Driving Score penalizes a collision and a stalled route alike, and trading one for the other leaves it nearly unchanged, whereas Success Rate depends on whether a route fails at all, and the trade removes far more failures than it creates.

\noindent\emph{Implication for evaluation.} A safety guard being penalized by the dominant metric reflects a limitation of the metric, rather than a weakness of the guard. Because existing benchmarks reward progress along the route, a vehicle that crashes after travelling farther can outscore one that waited for a gap that never opened. Our results give this a number: the same intervention reads as a $0.6\%$ improvement under DS and a $15.0\%$ improvement under SR. We therefore report the harmonic mean throughout, and suggest that long-tail benchmarks treat a cautious stop and a collision as categorically different outcomes, not differently weighted infractions.

\begin{figure}[tb]
    \centering
    \includegraphics[width=0.88\columnwidth]{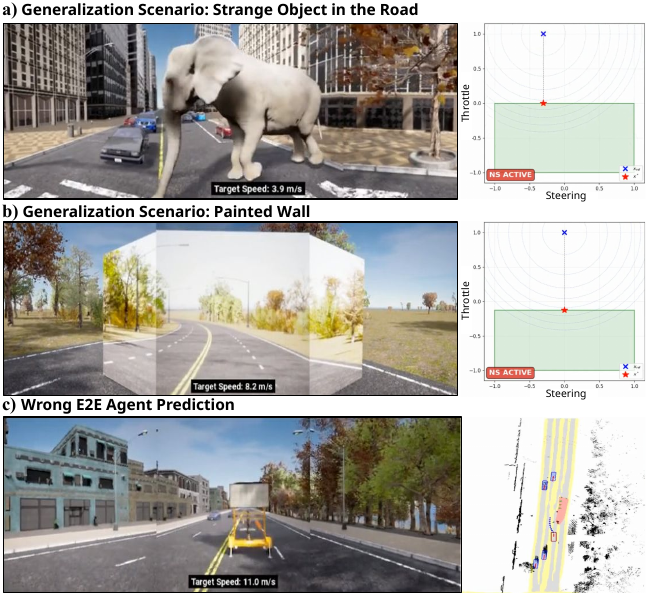}
    \caption{Guard successes (a) and (b) and limitations (c). Each right panel plots the nominal command (blue cross), the corrected command (red star), and the feasible set (green).}
    \label{fig:qualitative_cases}
\end{figure}

\noindent\emph{Qualitative reach and limit.} Figure~\ref{fig:qualitative_cases} shows both regimes directly. In (a) and (b) the policy accelerates toward an unfamiliar object and into a painted wall it reads as open road. These are exactly the failures the guard should catch: the obstacle is physically present even when the network does not recognize what it is, and the class-agnostic radar channel fires where the camera channel does not. Case (c) marks the limit. When the planner routes into the oncoming lane, no allowed command remains and the guard stops the car, taking a timeout instead of a collision. This is where the added timeouts come from: the guard can block an unsafe command, but it cannot invent a safe trajectory the planner never proposed, and it acts only on what perception reports.

\section{Conclusion}

A neuro-symbolic safety guard lets a trained driver keep everything it learned while obeying what physics and traffic law require. Attached at the final command interface, it turns what the agent sees into auditable limits on throttle, brake and steering. On the Fail2Drive long tail it raises Success Rate by $15\%$ and cuts safety-critical collisions by up to $53\%$ without lowering Driving Score. It attains the highest Harmonic Mean of any learned agent, and loses the least when the scene becomes unfamiliar. Because the guard reads only signals a modern stack already exposes and never alters its weights, the same construction attaches to any end-to-end agent without retraining. These results argue for judging an agent not by the distance it covers but by the conditions its commands are guaranteed to satisfy.

\bibliography{cybersar}

\end{document}